# Investigating the Perception of the Future in GPT-3, -3.5 and GPT-4

Diana Kozachek
Freie University of Berlin, Institut Futur
kozachekdiana@gmail.com

## ABSTRACT

This study explores the potential of GPT-3, GPT-3.5, and GPT-4, in generating human-like future scenarios to investigate each model's ability to perceive time. The methodology combines a coding-based experiment and an expert survey. The investigation involves fine- and prompt-tuning GPT-3, prompt-tuning GPT-3.5, and few-shot prompting GPT-4 with human-made future scenarios. The models and output are quantitatively and qualitatively analyzed. The survey invited practitioners from fields of foresight and futurology, AI, and NLP to assess whether differences in output can be identified. This study's findings suggest that GPT-3 and GPT-4 generated scenarios are difficult to distinguish from human-made ones, while GPT-3.5 performed more poorly. Yet none of the models can differentiate time horizons and their respective effects on the future from each other. And while no one knows the shape of things to come, this lack of understanding of a core concept of life invites future investigations.

## 1 INTRODUCTION

In March 2023, OpenAI released its so far most capable language model update, GPT-4, following the GPT-3.5 in January 2023, while their predecessor, GPT-3, launched in 2020, has already been a subject of research. Large Language Models (LLMs) can generate texts in various styles, languages, and lengths, as well as interacting with humans. This technology relies on prompts, allowing users to interact using natural language rather than code, making it accessible without technical expertise.

The methods of scenario development are regarded as the tools par excellence for future research [4]. Scenarios of possible futures and the description of the path that leads to these offer a basis for decision-making and at the same time open space for speculation. Scenic, creative texts make futures tellable, desirable or warn against them. Sohail Inayatullah, the UNESCO Chair in Futures Studies at the Sejahtera Center for Sustainability and Humanity, used ChatGPT, a model based on GPT-3.5 (now also GPT-4) specializing in conversations with humans, to construct and deconstruct futures. In his demonstrations, he asked ChatGPT about the Causal Layered Analysis (CLA) and the Change Progression Scenario Method (CPS) [5, 6]. He attested ChatGPT the potential to make work easier for futurologists. Spaniol and Rowland published a paper on ChatGPT a month later. Using their scenario checklist, they tested whether the output generated by the chatbot could be considered as future scenarios [13]. With these early-stage tests on ChatGPT, this study explores whether three variations of the foundation model can contribute a writing assistance to the scenario methodology, if they explicitly learn from human-made futures.

### 1.1 Research question and hypotheses

The research question is: Can GPT-3, GPT-3.5 or GPT-4 generate human-like future scenarios? The subject of research are a human-made scenario data set of 1080 scenarios as a benchmark and three types of machine-generated data sets: 1080 scenarios generated by fine- and prompt-tuned GPT-3 davinci, 1080 scenarios by prompt-tuned GPT-3.5 text-davinci-003, and 1080 scenarios generated by few-shot prompted GPT-4. The hypothesis and respective null hypothesis are: H1) Domain experts cannot distinguish GPT-3, GPT-3.5 or GPT-4 scenarios from human-made scenarios. H0) Domain experts recognize differences between human-made and GPT-generated scenarios.

### 1.2 Prerequisite: The human-made scenario data set

Scenario reports from public institutions were scraped in a partially automated web search. When selecting the content of the scenarios, the credibility of the authors, transparency regarding the client, and relevancy in terms of the contexts' breadth were quality standards, resulting in 238 reports. The search query was conducted using a Python script, through the search string: "future scenarios on European <keyword> in <year> site:.edu OR site:.gov OR site:.org OR site:.eu OR site:.mil after:2015" scenarios for the years 2030, 2035, 2040, 2045 and 2050 were requested combined with 12 keywords.

The following steps form the iterative data-processing after the step 1., Search, were: 2. Download, 3. Parsing, 4. Summarizing, 5. Cleaning, 6. Keyword Extraction, 7. Keyword Analysis, 8. Randomization, 9. Data Set Reduction [7]. If step 7. resulted in an imbalance regarding the time horizon, keywords or authors, the process was repeated with adapted search queries. Since future scenarios range in tonality, style, and goal, balancing the sources was crucial to achieve a suitable training data set for the fine-tuning and prompt-tuning with example scenarios.

**Table 1: Global Prompt-Parameter Settings**

| Prompt | Examples | Parameter | Global Setting |
|---|---|---|---|
| Write a future scenario on <keyword> in Europe in <year> as one sentence. The scenario should look like these examples: | 1: Scenario 1<br>2: Scenario 2<br>3: Scenario 3 | max_tokens<br>top_p<br>presence_penalty<br>frequence_penalty<br>temperature | 89<br>0.5<br>0.9<br>0.9<br>0.8 |

[a] This prompt was utilized to automate scenario generation in Python.

## 2 EXPERIMENTAL PROCESS

GPT-3.5 text-davinci-003 was prompt-tuned by testing 268 prompt-parameter variations [8]. The learnings from this prompting and prompt-tuning are used, in addition to the data set of human-generated scenarios from the first step, to fine-tune GPT-3 davinci [9]. Then, GPT-4 via ChatGPT Plus was utilized by few-shot prompting with human-made scenario examples. All data sets are evaluated through model-specific metrics.

### 2.1 Prompt-Tuning GPT-3.5 text-davinci-003

By prompt-tuning, the user can control the output through optimized natural language prompts or change five probability-based parameter settings: **max_tokens**, the combined length of the prompt and output, **top_p**, the probability mass, **presence_penalty**, avoiding same tokens within the output, **frequence_penalty**, returning the value of present tokens, and **temperature**, controlling the randomness of the output.

Prompt-tuned models can produce output that approach or even exceed the quality of the output of fine-tuned models [14]. Thirty variations of prompts were tested to find the most suitable output [10]. Then, 268 parameter-prompt variations were tested and evaluated by three experts, resulting in a global parameter setting with an inter-rater reliability of 100 % – these settings were applied to generate 1080 scenarios. The requirements resulted from the analysis of human-made scenarios. Table 1 shows a three-shot prompting with human-made scenarios.

The average number of words and characters determined the number of tokens. Each time horizon was requested 30 times for the 12 keywords, resulting in 1080 GPT-3.5 text-davinci-003-generated scenarios, e.g. for the keyword "education" and time horizon "2040", the examples would be: "Education is entirely in private hands, training students for the requirements of work in their client industries."/ "Students learn open-access from the best professionals in the world via virtual distance learning."/ "New starting points for easier access to higher education in Europe improved the quality of the study."

### 2.2 Fine-Tuning GPT-3 davinci

As the models under GPT-3 tend to generate undesirable results, fine-tuning was required. Using the human-made scenario data set, three GPT-3 davinci models were fine-tuned and evaluated in Table 2.

The model evaluation for the two best-performing fine-tunings, 1 and 3, can be seen in Figure 1. The **training_ token_accuracy** describes a model's ability to predict individual next tokens, when comparing the input data with the output data. A desirable result is an accuracy that grows throughout the training. The **training_loss** is the difference between the model's predictions and the output on the training data, indicating how well the model is fitting the training data. The model that trained over 6 epochs performed better during the fine-tuning. However, those results analyze the learning process, and the actual output needs to be investigated. Both hyperparameter settings were utilized to generate 1080 GPT-3 davinci scenarios.

### 2.3 Few-Shot Prompting GPT-4 via ChatGPT Plus with System Messages

Research suggests that GPT models perform better on a variety of tasks when given a natural language prompt containing 1-5 examples of preferred output. This approach is called few-shot prompting, allowing models to gain an understanding of a task and the context [1, 12].

GPT-3.5 text-davinci-003 displayed advanced abilities in passing a standardized legal exam, if presented with multiple-choices in prompt [2], indicating that the usage of a small sets of examples could also guide GPT-4 towards desired output. Another prompt optimization method can be achieved by using system messages. System messages help to establish a context and goal through instructions. Table 3 illustrates the differences in few-shot prompts. In method 3., GPT-4 focuses on one key factor only instead of enumeration all factors extracted from the examples as in 2. And while the differences are subtle, system messages are still at an early stage and require further trial and error to gain a full understanding of the underlying rules. Method 3 was applied to generate another data set of 1080 GPT-4 scenarios.

## 3 E-DELPHI SURVEY RESULTS

Twenty randomized samples from the five data sets (two versions of GPT-3 fine-tunings) were evaluated blindly by 42 experts (Table 4), and generally misclassified with varying normalized average ratings per scenario source, going as far as regarding a models' output that was trained on human-made examples as the most human-like, while human-made scenarios have the lowest average ratings across all data sets for human-likeness [11].

**Table 2: GPT-3 davinci output after three rounds of fine-tuning**

| Fine-Tuning Parameters | Prompt | Output |
|---|---|---|
| 1. epochs: 6; batch size: 2; learning_rate_ multiplyer: 0.1; prompt_loss_weight: 0.1 | Write a future scenario on education in Europe in 2030. The scenario should look like these examples: (see 2.1) | Virtual reality and augmented reality revolutionize the learning experience, provi-ding immersive and interactive educational content. |
| 2. epochs: 4; batch size: 2; learning_rate_ multiplyer: 0.5; prompt_loss_weight: 0.1 | Write a future scenario on education in Europe in 2030. The scenario should look like these examples: (see 2.1) | Cross-border education programs flourish, connecting European students and teachers across the borders of Europe to study. |
| 3. epochs: 4; batch size: 2; learning_rate_ multiplyer: 0.1; prompt_loss_weight: 0.1 | Write a future scenario on education in Europe in 2030. The scenario should look like these examples: (see 2.1) | In the year 2030, education in Europe adopted virtual classrooms, allowing students to learn from teachers around the world. |

[a] Prompted with the Parameter Settings from Table 1; Examples from 2.1

**Table 3: GPT-4 with Zero vs Three-Shot Prompts and System Messages**

| Method | Prompt | Output |
|---|---|---|
| 1. Zero-Shot-Prompt | Write a future scenario on education in Europe in 2030. | European education blends virtual and physical learning through online platforms. |
| 2. Three-Shot-Prompt | Write a future scenario on education in Europe in 2030. The scenario should look like these examples: (see 2.1) | By 2030, European education embraces a hybrid model that fosters global collabo-ration, leveraging online platforms and in-person interactions, connecting students, teachers, and professionals, and industry, for a dynamic, personalized learning experi-ences. |
| 3. Three-Shot-Prompt & System Messages | Write a future scenario on education in Europe in 2030. The scenario should look like these examples: (see 2.1) The tonality is informative, goal-oriented and engaging, focusing on one single key factor per scenario. | By 2030, European education has been revolutionized by the widespread adoption of virtual classrooms and distance learning platforms, allowing students to learn from experts around the world in an interactive way. |

**Table 4: Confusion Matrix for the Classification of Machine-Generated or Human-Made Scenarios**

| Source | Human-made Data | Machine-generated Data |
|---|---|---|
| Categorized as human-made | Correctly categorized as human-made: 11.17 | Falsely categorized; GPT-3: 19.25 (6 & 4 epochs)<br>Falsely categorized; GPT-3.5: 18.25<br>Falsely categorized; GPT-4: 19.17 |
| Categorized as machine-generated | Falsely categorized as Human-made: 21.00 | Correctly categorized; GPT-3: 14.00(6 & 4 epochs)<br>Correctly categorized; GPT-3.5: 15.00 |

## 4 DISCUSSION

An overview of the scenarios regarding the intrinsic metrics of average scenario length, the diversity metric of unique words and the quality metric of incompleteness regarding the output was conducted in comparison to the benchmark (human-made scenarios). Additionally, the overlap of n-grams gives away the ratio of similarities in data sets, where n-1 stands of single word similarity, n-2 for word pair similarity and n-3 for combinations of three words that can be found in the GPT-generated and human-made scenario data sets. Table 5 gives a broad impression of the differences.

Revisiting the hypothesis, out of all three models' outputs, GPT-3.5 performed the worst amongst experts' ratings, displays a low lexical diversity and a high number of incomplete output scenarios, while still being rated as more human-like than human-made scenarios. GPT-3 (epochs 4), trained on human-made scenarios, on the other hand, was rated as the most human-like and displayed a broad diversity – but was also the most resource intense to set up. GPT-4, despite having the lowest lexical diversity and a low n-gram overlap with the human-made data, performed in the middle field, when it comes to the experts' ratings. H1 can be accepted, as the survey showed that experts are unable to differentiate human-made scenarios from machine-generated ones.

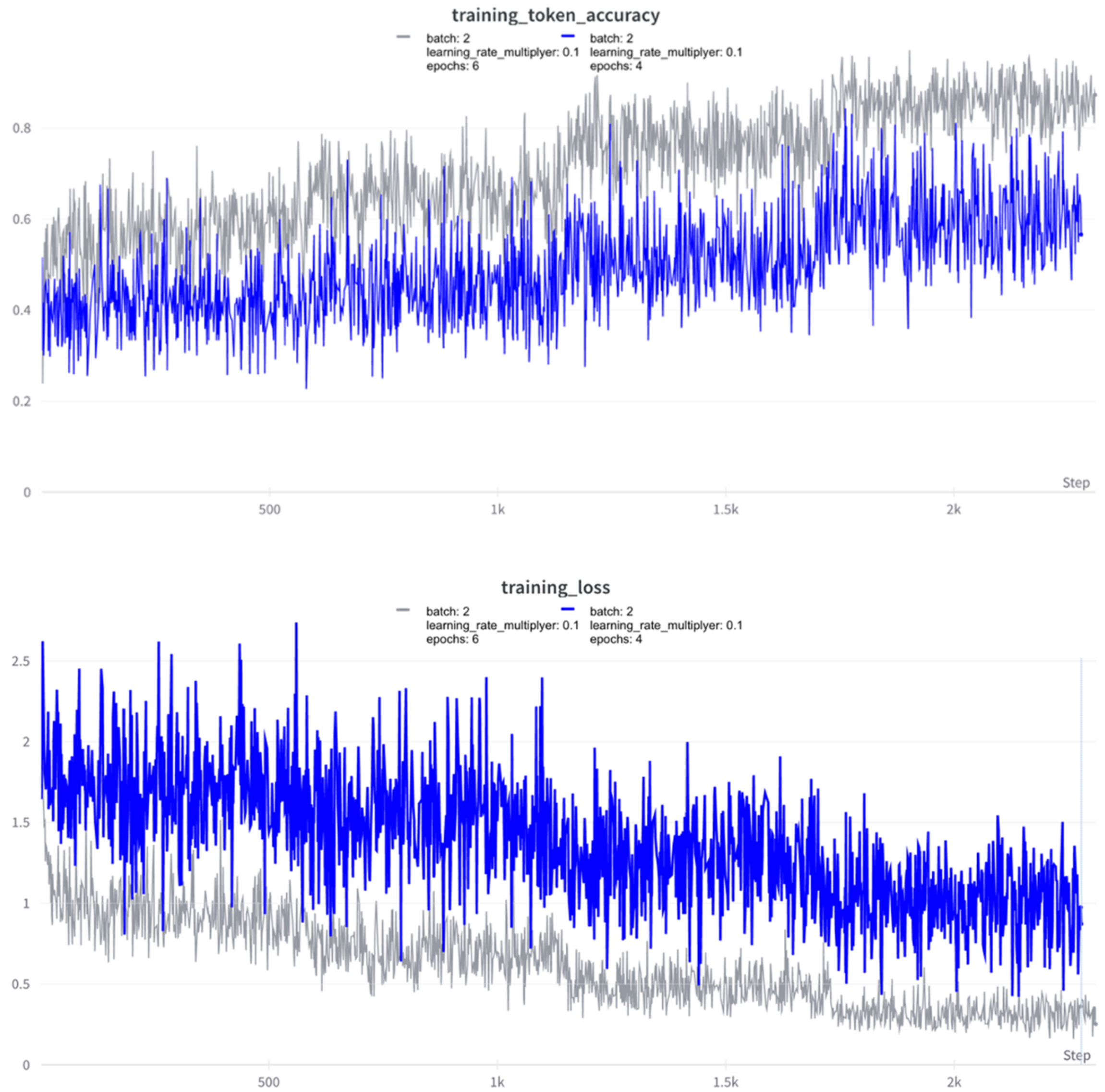

**Figure 1: The training_token_accuracy and training_loss in two rounds of fine-tuning GPT-3 davinci with different hyperparameters: training_token_accuracy for 6 epochs (grey line) ends at 0.872, the training_loss ends at 0.252 at training step 2310. training_token_accuracy for 4 epochs (blue line) ends at 0.566, the training_loss ends at 0.812 at training step 2310.**

**Table 5: Quantitative Data Set Comparison**

| Source | Incomplete Scenarios | Average Length | Unique words per data set | N-Gram Overlap: n1; n2; n3 |
|---|---|---|---|---|
| Human | (Benchmark) | 148 characters | 4002 | (Benchmark) |
| GPT-3, epoch: 6 | 418 | 104 characters | 1820 | 0.8; 0.4; 01 |
| GPT-3, epoch: 4 | 252 | 162 characters | 2834 | 0.6; 0.2; 0.0 |
| GPT-3.5 | 106 | 135 characters | 2331 | 0.6; 0.2; 0.0 |
| GPT-4 | 31 | 128 characters | 2261 | 0.6; 0.2; 0.0 |

**Table 6: GPT-4 repeating itself mostly within different Time Horizons**

| Model | Prompt | Output |
|---|---|---|
| GPT-4 | Write a future scenario on ecology in Europe in **2030**. | By 2030, **renewable hydrogen has become a** mainstream **energy carrier,** significantly contributing to the **global energy** transition. |
| GPT-4 | Write a future scenario on ecology in Europe in **2040**. | By 2040, **renewable hydrogen has become a** vital **energy carrier** in the **global** quest for sustainability. |
| GPT-4 | Write a future scenario on ecology in Europe in **2050**. | By 2050, **renewable hydrogen has become** an integral part of the **global energy** landscape a versatile and reliable **energy carrier.** |

**Table 7: GPT-3-5 text-davinci-003 repeating itself within different Time Horizons**

| Model | Prompt | Output |
|---|---|---|
| GPT-3.5 text-davinci-003 | Write a future scenario on ecology in Europe in **2030**. | The adoption of renewable hydrogen as an energy carrier has transformed industries such as transportation and energy, leading to a more sustainable future. |
| GPT-3.5 text-davinci-003 | Write a future scenario on ecology in Europe in **2040**. | The adoption of renewable hydrogen as an energy carrier has transformed industries such as transportation and energy, leading to a more sustainable future. |
| GPT-3.5 text-davinci-003 | Write a future scenario on ecology in Europe in **2050**. | The adoption of renewable hydrogen as an energy carrier has transformed industries such as transportation and energy, leading to a more sustainable future. |

One issue, however, applies to all the models: The lack of differentiation between time-horizons. None of the models was able to differ between 2030, 2040 and 2050, while generating output within the same session, and therefore, being able to recall the full output. Especially the prompt-tuned models, GPT-3 and GPT-3.5 should be able to avoid repetition through parameter settings, but ended up repeating themselves, nevertheless. Table 6 shows examples of GPT-4 repeating itself, Table 7 provides more severe examples for GPT-3.5.

## 5 CONCLUSION

While humans experience time on multiple dimensions, measurable scales can play a secondary role. Physical time manifests itself in real world situations. When it comes to thinking strategically about the future, however, humans draw from their experiences with abstract concepts of processes, such as (de-)growth, development, or aging. In fact, new evidence of a predictive coding hierarchy in the human brain when listening to speech indicated that we draw distant predictions by just listening, while LLMs operate on a token-to-token basis. Complex predictions occur over multiple timescales and levels of representations across the cortical hierarchy [3]. Comparing the coding hierarchy in the brain to language models, the sheer ability of forecasting becomes inherently human, unfolding over a multi-dimensional space beyond clocks, calendars, or Big-O-notations.

The factor of experience gives humans a subjective timeline that cannot be computed (yet) but becomes crucial when it comes to futurology. When experts develop futures, they apply domain expertise, stemming from experiences in the real world. Knowledge about depth and speed of progress over time in a specific field are human advantages that apparently did not transfer into the training data base of any GPT model – the low n-gram overlap indicates only few similarities. The lack of "thinking" further after a specific scenario for 2030 has been generated does not exists, neither. And as the holistic understanding of factors that influence progress within a field require reasoning, instead of a database only, none of the models offer a reasonable solution for the scenario generation. A deep dive into the subjective and objective understanding of time in language models would require studies across multiple disciplines. Futurology and Foresight, however, should be aware of this factor when considering utilizing GPT models.